\documentclass[journal]{IEEEtran}
\usepackage{cite}
\usepackage{xcolor}

\ifCLASSINFOpdf
  \usepackage[pdftex]{graphicx}
  \graphicspath{ {./} }
\else
  \usepackage[dvips]{graphicx}
\fi
\usepackage{xcolor}

\usepackage{amsmath,amssymb}
\usepackage{algorithm,algpseudocode}
\usepackage{url}
\newcommand{\etal}{\textit{et al}.}
\usepackage{balance}
\usepackage{setspace}
\usepackage{flushend}

\begin{document}
%

\title{Adversarial Attack on Skeleton-based Human Action Recognition}

%
%
%

\author{Jian~Liu,
        Naveed~Akhtar,
        and~Ajmal~Mian,
}

\maketitle

\begin{abstract}
Deep learning models achieve impressive performance for skeleton-based human action recognition. However, the robustness of these models to adversarial attacks remains largely unexplored due to their complex spatio-temporal nature that must represent sparse and  discrete skeleton joints. 
This work presents the first adversarial attack on skeleton-based action recognition with  graph convolutional networks.
The proposed targeted attack, termed Constrained Iterative Attack for Skeleton Actions (CIASA), perturbs joint locations in an action sequence such that the resulting adversarial sequence preserves the temporal coherence, spatial integrity, and the anthropomorphic plausibility of the skeletons.
CIASA achieves this feat by satisfying  multiple physical constraints, and employing spatial skeleton realignments for the perturbed skeletons along with regularization of the adversarial skeletons with Generative networks. We also explore the possibility of semantically imperceptible localized attacks with CIASA, and succeed in fooling the state-of-the-art skeleton action recognition models with high confidence. CIASA perturbations show high transferability for black-box attacks. We also show that the perturbed skeleton sequences are able to induce adversarial behavior in the RGB videos created with computer graphics.    
A comprehensive evaluation with NTU and Kinetics datasets ascertains the effectiveness of CIASA for graph-based skeleton action recognition and reveals the imminent threat to the spatio-temporal deep learning tasks in general.

\end{abstract}


%
\IEEEpeerreviewmaketitle

\section{Introduction}
\label{sec:Intro}
%
%
%
%

\begin{figure}[t]
\centering
\includegraphics[width=0.45\textwidth]{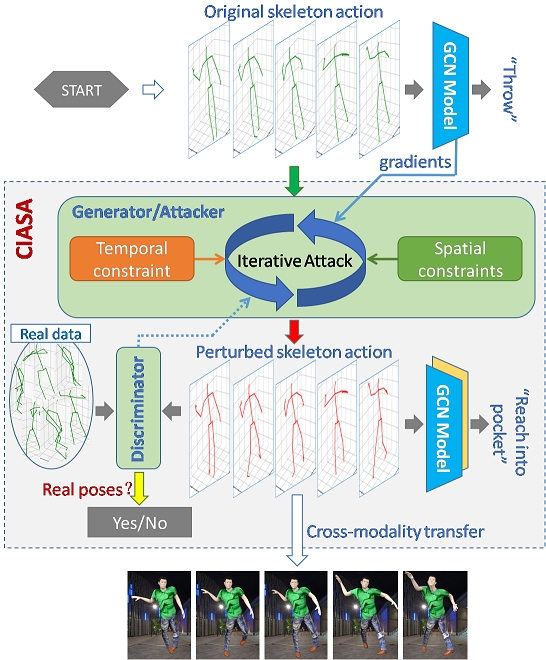}
\caption{Constrained Iterative Attack for Skeleton Actions (CIASA) schematics. Model gradients are computed for input action sequence to iteratively minimise the model's loss for a target label in small step, while accounting for the relevant spatio-temporal constraints. A generator-discriminator framework further ensures anthropomorphic plausibility of the sekeltons. Besides cross-model transferability, the attack can also affect RGB videos generated with computer graphics using the skeletons perturbed by CIASA.}  
\label{fig:teaser}
\end{figure}

\noindent Skeleton representation provides the  advantage of capturing accurate human pose information  while being invariant to action-irrelevant details  such as scene background, clothing patterns and illumination conditions. This makes skeleton-based action recognition an appealing approach~\cite{zanfir2013moving,devanne20153,du2015skeleton,vemulapalli2016rolling,ke2017new,yang2017latent}. The problem is also interesting for multiple application domains, including security, surveillance, animation and human-computer interactions etc. Recent contributions in this direction predominantly exploit deep models to encode spatio-temporal dependencies of the skeleton sequences~\cite{liu2017skeleton,yan2018spatial,cao2018body,shi2019two}, and achieve remarkable recognition accuracy on benchmark action datasets~\cite{shahroudy2016ntu,yun2012two,ofli2013berkeley,hu2015jointly}.

Although deep learning has been successfully applied to many complex problems, it is now known that deep models are vulnerable to adversarial attacks~\cite{szegedy2013intriguing,goodfellow2014explaining}. These attacks can alter model predictions at will by adding imperceptible perturbations to the input. After the discovery of this intriguing weakness of deep learning~\cite{szegedy2013intriguing}, many adversarial attacks have surfaced for a variety of vision tasks~\cite{papernot2016limitations,su2019one,mirjalili2017soft,lin2017tactics}. Developing and investigating these attacks not only enhances our understanding of the inner workings of the neural networks~\cite{LUTA}, but also provides valuable insights for improving the robustness of deep learning in practical adversarial settings.

Deep models for skeleton-based action recognition may also be vulnerable to adversarial attacks. However, adversarial attacks on these models are yet to be explored. A major challenge in this regard is that the skeleton data representation differs significantly from image representation, for which the existing attacks are primarily designed.
Human skeleton data is sparse and discrete that evolves over time in rigid spatial configurations. This prevents an attacker from freely modifying the skeletons without raising obvious attack  suspicions. Skeleton actions also allow only  subtle perturbations along the temporal dimension to preserve the natural action dynamics. 
In summary, adversarial attacks on skeleton data must carefully account for the skeleton's spatial integrity, temporal coherence and anthropomorphic plausibility. Otherwise, the attack may be easily detectable. These challenges have so far kept skeleton-based action recognition models away from being scrutinized for adversarial robustness.


In this work, we present the {\em first} adversarial attack on deep skeleton action recognition. In particular, we attack the (most) promising branch of graph convolutional networks~\cite{kipf2016semi} for   skeleton-based action  recognition~\cite{yan2018spatial}. 
These models represent actions as spatio-temporal graphs that encode intra-body and inter-frame connections as edges, and body joints as nodes. Graph convolution operations are  leveraged to model the spatio-temporal dependencies within the skeleton sequences. The physical significance of nodes and edges in these models  imposes unique constraints over the potential attacks. For instance, the graph nodes for a  skeleton sequence can not be added or removed because the number of joints in the skeleton must remain fixed. Similarly, the lengths of intra-body edges in the graph can not be altered arbitrarily as they represent bones. Moreover, inter-frame edges must always connect the same joints along the temporal dimension.
Rooted in the skeleton data, such constraints thoroughly distinguish the adversarial attacks on skeleton-based action recognition models from the attacks  developed for other kinds of graph networks~\cite{zugner2018adversarial}.

We develop an iterative scheme called Constrained Iterative Attack for Skeleton Actions (CIASA), to generate the desired adversarial skeleton sequences, see Fig.~\ref{fig:teaser}.  
For a given action, CIASA iteratively perturbs its skeleton sequence in small steps to minimize the model prediction loss for a pre-selected target class while satisfying multiple physical constraints to keep the resulting adversarial sequence natural.   
In particular, it accounts for spatio-temporal constraints that preserve intra-skeleton joint connections, inter-frame joint connections, and the skeleton bone lengths using a mechanism termed `spatial skeleton realignment'.
For perturbation imperceptibility, it restricts the $\ell_{\infty}$-norm of the added noise. Additionally, it  imposes external temporal dynamics constraints for imperceptible evolution of the adversarial patterns in the skeleton sequence.
To further ensure anthropomorphic plausibility of the adversarial skeleton sequence, it exploits the Generative Adversarial Network (GAN) framework~\cite{goodfellow2014generative}. The used GAN configuration reduces the difference between the distribution of  adversarial samples generated by our iterative scheme and the clean ground truth samples.  

We analyze the proposed attack by allowing different modes in which CIASA can be used by an attacker. Analogous to standard image based attacks, we allow perturbation of all skeleton joints in the \textit{basic} mode. In a \textit{localized} mode, we provide the flexibility of perturbing only localized regions, e.g.~legs of skeleton. This type of attack is particularly suitable to skeleton actions where an attacker may independently alter motion of the least relevant joints for an action to change the prediction. We also introduce an \textit{advanced} attack mode that further allows a hierarchical magnitude variation in joint perturbations based on the graph structure of the joints.

The notion of localized perturbation also leads to \textit{semantically} imperceptible perturbations under CIASA where significant perturbation still remains hard to perceive because it is applied to the least significant joints for the original action semantics. Besides demonstrating high fooling rates for the state-of-the-art graph skeleton action recognition model ST-GCN~\cite{yan2018spatial} on NTU~\cite{shahroudy2016ntu} and Kinetics~\cite{kay2017kinetics} datasets, we also show high cross-model transferability of the proposed attack. 
Additionally, we show that videos generated (using computer graphics) from the adversarial skeletons (CIASA's advanced mode) result in lower action recognition accuracy implying that the attack can be launched in the real world. 
To the best of our knowledge, this is the first of its kind demonstration of transferability of adversarial attacks beyond a single data modality.

The rest of this article is organized as follows.  We review the related literature in Section~\ref{sec:RW}. The relevant concepts of graph skeleton action recognition are revisited in Section~\ref{sec:BG} along with the problem formulation. In Section~\ref{sec:MD}, we give the implementation details of the proposed attack scheme. Experimental results are provided in Section~\ref{sec:EXP}. The article concludes in Section~\ref{sec:Conc}.

 

\section{Related Work}
\label{sec:RW}

\subsection{Skeleton-based Action Recognition}
The use of skeleton data in action recognition becomes popular as reliable skeleton data can be obtained from modern RGB-D sensors (e.g.~Microsoft Kinect), or extracted from images taken from a single RGB  camera~\cite{mehta2017vnect}. A  skeleton action is represented as a sequence of human skeletons, which encode rich spatio-temporal information regarding human motions. Early research in skeleton-based action recognition formulated skeleton joints and their temporal variations as trajectories~\cite{devanne20153}. Huang \etal~\cite{huang2016deep} incorporated the Lie group structure into the task, and transformed the high-dimensional Lie group trajectory into temporally aligned Lie group features for skeleton-based action recognition.

To leverage the power of convolutional neural network, Du \etal~\cite{du2015skeleton} represented a skeleton sequence as a matrix by concatenating the joint coordinates. The matrix is arranged as an image which can be fed into CNN for recognition. Similarly, Ke \etal~\cite{ke2017new} transformed a skeleton sequence into three clips of gray-scale images that encode spatial dependencies between the joints by inserting reference joints. To fit the target neural networks, these methods  re-size the transformed images. Liu \etal~\cite{liu2017skepxels} proposed a universal unit ``skepxel'' to create images of arbitrary dimensions for CNN processing. In addition to CNNs, Recurrent Neural Networks are also employed to model temporal dependencies in skeleton based human action analysis~\cite{du2015hierarchical,veeriah2015differential,shahroudy2016multimodal}.

To directly process the sparse skeleton data with neural networks, graph convolutional network (GCN)~\cite{kipf2016semi} is used for action recognition. Since GCN is particularly relevant to this work, we review its relevant literature and application to action recognition in more detail.

\subsection{Graph Convolution Networks}
The topology of human skeleton  joints is a typical graph structure, where the joints and bones are respectively interpreted as graph nodes and edges. Consequently, there have been several recent attempts in modeling human skeleton actions using graph representation and exploiting the spatio-temporal dependencies in skeleton sequences with the help of graph-based convolutional network (GCN).

Yan~\etal~\cite{yan2018spatial} used graph convolutional networks as a spatial-temporal model (ST-GCN) that aims to capture embedded patterns in the spatial configuration of skeleton joints and their temporal dynamics simultaneously. Along the skeleton sequence, they defined a graph convolution operation, where the input is the joint coordinate vectors on the graph nodes. The convolution kernel samples the neighboring joints within the skeleton frame as well as the temporally connected joints at a defined temporal range.

Tang~\etal~\cite{tang2018deep} incorporated deep reinforcement learning with graph neural network to recognize skeleton-based actions. Their model  distills the most informative skeleton frames and discards  the ambiguous ones. As opposed to previous works where joints dependency is limited in the real physical connection (intrinsic dependency), they proposed extrinsic joint dependency, which exploits the relationship between joints that have physical disconnection. Since, graph representation of skeleton is crucial to graph convolution, Gao~\etal~\cite{gao2018generalized} formulated the skeleton graph representation as an optimization problem, and proposed graph regression to statistically learn the underlying graph from multiple observations. The learned sparse graph pattern links both physical and non-physical edges of skeleton joints, along with the spatio-temporal dimension of the skeleton action sequences.

To justify the importance of bones' motions in skeleton action recognition, Zhang~\etal~\cite{zhang2018graph} focused on skeleton bones and extended the graph convolution from graph nodes to graph edges. Their proposed graph edge convolution defines a receptive field of edges, which consists of a center edge and its spatio-temporal neighbours. By combining the graph edge and node convolutions, they proposed a two-stream graph neural network, which achieved remarkable performances on benchmark datasets. Similarly, Shi~\etal~\cite{shi2018adaptive} also proposed a two-stream framework to model joints and bones information simultaneously. 


\subsection{Adversarial Attacks on Graph Data}

Adversarial attacks~\cite{szegedy2013intriguing}  have recently attracted significant research attention~\cite{akhtar2018threat}, resulting in few attacks on graph data as well. However, compared to the adversarial attacks for image data~\cite{goodfellow2014explaining,liu2016delving,papernot2017practical,xiao2018spatially}, several new challenges appear in attacking graph data~\cite{sun2018adversarial}. First, the graph structure and features of graph nodes are in discrete domain with certain pre-defined structures, which leaves a lower degree of freedom for creating adversarial perturbations. Second, the imperceptibility of adversarial perturbations in graph data is neither easy to define nor straightforward to achieve, as the discrete graph data inherently prevents infinitesimal small changes~\cite{zugner2018adversarial}.

Dai~\etal~\cite{dai2018adversarial} focused on attacking structural information, i.e. adding/deleting graph edges, to launch adversarial attacks on graph structured data. Given the gradient information of target classifier, one of their proposed attacks modifies the graph edges that are most likely to change the objective. 
In addition to modifying graph edges, Z{\"u}gner~\etal~\cite{zugner2018adversarial} adopted an attack strategy to modify the graph node features as well as graph edge structure. To ensure the imperceptibility of adversarial perturbations, they designed constraints based on power-law~\cite{bessi2015two} to preserve the degree distribution of graph structures and feature statistics. 

Being atypical graph data, human skeletons have several unique properties. In a human skeleton, the graph edges represent rigid human bones, which connect finite number of human joints to form a standard spatial configuration. Unlike graph data with mutable graph structure (e.g. social network graph~\cite{newman2002random}), the human bones are fixed in terms of both joint connections and bone lengths. This property implies that attacking human skeletons by adding or deleting bones will be detected easily by observers.
The hierarchical nature of human  skeleton data is also different from normal graph data, as in human skeleton the motion of children joints/bones are affected by their parents' behaviours. This chain-like motion kinetics of human skeletons must be considered when launching adversarial attacks on skeleton actions. Hence, despite the existence of adversarial attacks on graph data, robustness of skeleton based human action recognition against adversarial attacks remains largely unexplored. 

In this work, we specifically focus on adversarial attacks on human skeleton sequences to fool skeleton-based action recognition models. To design effective and meaningful attacks, we take the spatial and temporal attributes of skeleton data into account while creating the adversarial perturbations. Due to its wide-spread use in graph convolution network based action recognition, we select ST-GCN~\cite{yan2018spatial} as our target model, and launch our attack against it. However, our attack is generic for similar graph based model. In  the section to follow, we formulate our problem in the context of skeleton based human action recognition.

\section{Problem Formulation}
\label{sec:BG}
To formulate the problem, we first 
briefly revisit the spatio-temporal graph convolutional network ST-GCN~\cite{yan2018spatial} for skeleton-based action recognition. Using  this prerequisite knowledge, we subsequently formalize our problem of adversarial attacks on skeleton action recognition.

\subsection{Revisiting ST-GCN}
\label{sec:BG_Rev}
An action in skeleton domain is represented as a sequence of $T$ skeleton frames, where every skeleton consists of $N$ body joints. Given such $N \times T$ volumes of joints, an undirected spatio-temporal graph $G=(V,E)$ can be constructed, where $V$ denotes the node set of graph and $E$ is the edge set. Here, $V=\{v_{ti}|t=1,\dots,T, i=1,\dots,N\}$ encodes the  skeleton joints. An element  `$v$'  of this set can also be  considered to encode a joint's Cartesian coordinates.
Two kinds of graph edges $E$ are defined for joints, namely; intra-body edge $E^S$ and inter-frame edge $E^F$. Specifically, $E^S$ is represented as an $N\times N$ adjacency matrix of graph nodes, where the matrix element $E^S_{ij}=1 | i\neq j$ identifies that a physical bone connection exists between the body joint $v_i$ and $v_j$. The inter-frame edges $E^F$ denotes the connections of the same joints between consecutive frames, which can also be treated as temporal trajectories of the skeleton joints.

Given the spatio-temporal skeleton graph $G$, a graph convolution operation is defined by extending the conventional image-based convolution.
Along the spatial dimension,  graph convolution is conducted on a graph node $v_i$ around its neighboring nodes $v_j\in B(v_i)$:
\begin{equation}
    f_{out}(v_i) = \sum_{v_j \in B(v_i)}\frac{1}{Z_i(v_j)}f_{in}(v_j)\cdot w(l_i(v_j)),
\end{equation}
where $B$ is the sampling function to define a neighboring node set for the joint $v_i$, $f_{in}$ is the input feature map, $w$ is the weight function which indexes convolution weight vectors based on the labels of neighboring nodes $v_j$, and $Z_i(v_j)$ is the number of neighboring nodes to normalize the inner product. The labels of neighboring nodes are assigned with a labelling function $l:B(v_i)\rightarrow \{0,\dots, K-1\}$, where $K$ defines the spatial kernel size. ST-GCN employs different skeleton partitioning strategies for the labelling purpose.
To conduct graph convolution in spatio-temporal dimensions, the sampling function $B(v)$ and the labelling function $l(v)$ are extended to cover a pre-defined temporal range $\Gamma$, which decides the temporal kernel size.

ST-GCN~\cite{yan2018spatial} adopts the implementation of graph convolution network in~\cite{kipf2016semi} to create a 9-layer neural network with temporal kernel size $\Gamma=9$ for each layer. Starting from 64, the number of channels is doubled for every 3 layers. The resulting tensor is pooled at the last layer to produce a feature vector $f_{final}\in \mathbb{R}^{256}$, which is fed to a Softmax classifier for predicting the action label. The network mapping function is compactly represented as:
\begin{equation}
    Z_{G,c} =\mathcal{F}_{\theta}(V,E) = \mathop{\arg\max} <\mathrm{softmax}(f_{final})>,
\end{equation}
where 
$\theta$ denotes the network parameters. We use $Z_{G,c}$ to denote the probability of assigning spatio-temporal skeleton graph $G$ to class $c\in C=\{1,2,\dots,c_k\}$. After training, the network  parameters are fine-tuned to minimize the cross entropy loss between the predicted class $c$ and the ground truth $c_{gt}$ that maximizes the probability $Z_{G,c}|c=c_{gt}$ for the dataset under consideration.

\subsection{Adversarial Attack on Skeleton Action Recognition}
Given an original spatio-temporal skeleton graph $G^0=(V^0,E^0)$, and a trained ST-GCN model $\mathcal{F}_{\theta}$, our goal is to apply adversarial perturbation to the graph $G^0$, resulting in a perturbed graph $G{'}=(V{'},E{'})$ that satisfies  the following broad constraint:
\begin{equation}
\label{func:aa}
Z_{G{'},c}=\mathcal{F}_{\theta}(V',E'), ~\textrm{s.t.}~c\neq c_{gt}
\end{equation}
Below, we examine this objective from various aspects to compute effective adversarial perturbations for the skeleton action recognition.

\subsubsection{Feature and structure perturbations }
\label{sec:BG_ao}

As explained in Section~\ref{sec:BG_Rev}, $V$ denotes  the skeleton joints whose elements can be represented as the Cartesian coordinates of  joints, e.g.~$v_{ti}:\{x_{ti},y_{ti},z_{ti}\}$. For a particular node $v_{ti}$ in the skeleton graph $G$, an adversarial attack can change its original location such that $v{'}_{ti}=v^0_{ti}+\rho_{ti}$, where $\rho_{ti}\in \mathbb{R}^3$ is the adversarial perturbation for the node $v_{ti}$. We refer to this type of perturbation as \textit{feature perturbation}. Alternatively, one can define \textit{structure perturbation} that aims at changing the adjacency relationship in a graph such that $E{'}_{ij}\neq E^0_{ij}|i,j\in\mathcal{V}$, where $\mathcal{V}$ denotes the set of affected graph nodes.

In a spatio-temporal skeleton graph $G$, perturbing  edges have strong physical implications. Recall that intra-body connections of joints define the rigid bones within a skeleton, and inter-frame connections define the temporal movements of the joints. Changes to these connections can lead to skeleton sequences that cannot be interpreted as any  meaningful human action. 
Hence, the objective in Eq.~\ref{func:aa} must further be constrained to preserve the graph structure while computing the perturbation. To account for that, we must modify the overall constraint to:
\begin{equation}
\label{func:ab}
Z_{G{'},c}=\mathcal{F}_{\theta}(V',E^0), ~\textrm{s.t.}~c\neq c_{gt}.
\end{equation}

\subsubsection{ Perturbation imperceptibility}
\label{sec:BG_imp}

Imperceptibility is an important attribute of adversarial attacks, as adversaries are likely to fool deep models in unnoticeable ways. Here, we explore perturbation imperceptibility in the context of skeleton actions. This leads  to further constraints that must be satisfied when launching adversarial attacks on a skeleton graph $G$.

For the conventional image data, imperceptibility of perturbations is typically achieved by restricting  $\|\rho\|_p < \xi$, where $\|\cdot\|_p$ denotes the $\ell_p$-norm of a vector with $p\in [0,\infty)$, and $\xi$ is a pre-defined constant~\cite{akhtar2018defense}. For the skeleton graph data, however, the graph structure is discrete and graph nodes are dependent on each other, which makes it more challenging to keep a valid perturbation fully imperceptible. We  tackle the challenge of perceptibility for  skeleton perturbations from multiple point of views that results in multiple constraints for the overall objective, as explained in the following paragraphs. 

\vspace{2mm}
\noindent{\textbf{Joints variation constraint:}}

Focusing on \textit{feature perturbation} on skeleton graph, the location of a target skeleton joint is changed such that $v{'}_{ti}=v^0_{ti}+\rho_{ti}$.
It is intuitive to constrain $\rho$ of every target joint in a small range to avoid breaking the spatial integrity of the skeleton. Hence, we employ the following constraint:
\begin{equation}
    \|\rho_{ti}\|_{\infty}\leqslant \epsilon_{i}~|~ t\in[1,\dots,T];i\in[1,\dots,N],
\end{equation}
where $\|\cdot\|_{\infty}$ denotes $\ell_{\infty}$-norm, and $\epsilon_{i}$ is a pre-fixed constant. By restricting the joint variations to a small $\ell_{\infty}$-ball,
we encourage perturbation imperceptibility.  
From the implementation view point, when the ball radius
$\epsilon$ is constant for all joints, we call it \textit{global clipping} of the perturbed joints, and when the value of $\epsilon_{i}$ is joint-dependent, we call it \textit{hierarchical clipping}.

\vspace{2mm}
\noindent{\textbf{Bone length constraint:}}

In a skeleton graph $G$, the intra-body graph connections $E^S$ represent rigid human bones, hence their lengths must be preserved despite the perturbations. In the case of $E^S_{ij}=1|i\neq j$, the length of the bone between joint $i$ and $j$ at frame $t$ can be calculated as $B_{ij,t}=\|v_{ti}-v_{tj}\|_2$. After applying  perturbations to the graph, the new bone length ${B'}_{ij,t}=\|v'_{ti}-v'_{tj}\|_2$ should satisfy the following:
\begin{equation}
\label{func:bl}
    B_{ij,t}={B'}_{ij,t}~|~ t\in[1,\dots,T] ~\textrm{s.t.}~ E^S_{ij}=1.
\end{equation}

\noindent{\textbf{Temporal dynamics constraint:}}
Due to the spatio-temporal nature of skeleton action graphs, we disentangle the restrictions over perturbations into spatial and temporal constrains. Previous paragraphs mainly focused on the spatial constraints. Here, we analyze the problem from a temporal perspective.

A skeleton action is a sequence of skeleton frames that transit smoothly along the temporal dimension. A skeleton perturbation may lead to random jitters in the temporal trajectories of the joints  and compromise the smooth temporal dynamics of the target skeleton action. To address this problem, we impose an explicit temporal constraint over  the perturbations. Inspired by~\cite{mehta2017vnect}, we penalize acceleration of the perturbed joints to enforce temporal stability. Given consecutive perturbed skeleton frames $f'_{t-1}$,$f'_t$, and $f'_{t+1}$, the acceleration is calculated as $\ddot{f'_t}=f'_{t+1}+f'_{t-1}-2f'_t$. Note that, $f'_t=\{v'_{ti}|i=1,\dots,N\}$, where $N$ is the number of perturbed skeleton joints. The calculation of acceleration is conducted on individual joints. We optimize our attacker, $\mathcal{A}$ (discussed further below) by including the following  temporal smoothness loss in the overall objective:
\begin{equation}
\label{func:loss_smooth}
    \mathcal{L}_{smooth}(\mathcal{A})=\frac{1}{T-1}\sum^T_{t=2}\ddot{f'_t}=\frac{1}{T-1}\sum^T_{t=2}\sum^N_{i=1}\ddot{v'_{ti}},
\end{equation}
where $T$ denotes the number of time steps considered. In the text to follow, we use $\ddot{f'_t}$ to denote the joint acceleration for notational simplification.

\subsubsection{Anthropomorphic  plausibility}
\label{sec:Ant}

After adversarial perturbation is applied to a skeleton, the resulting skeleton can become anthropomorphically implausible. For instance, the perturbed arms and legs may bend unnaturally, or significant self-intersections may occur within the perturbed armature. Such unnatural behaviour can easily raise attack suspicions. Therefore, this potential behavior needs to be regularized while computing the  perturbations. 

Let $\mathcal{P}$ define the distribution of natural skeleton graphs. A  sample graph $G^0$ is drawn from this distribution with probability $\mathcal{P}(G^0)$. We can treat an adversarial skeleton's graph $G'$ to be a sample of another similar distribution $\mathcal{P'}$. The latter distribution should closely resemble the former under the restriction of minimal perturbation of joints and anthropomorphic plausibility of the skeletons. 
Hence, to obtain effective adversarial skeletons we aim at reducing the distribution gap between $\mathcal{P}$ and $\mathcal{P'}$. To that end, we employ a Generative Adversarial Network  (GAN)~\cite{goodfellow2014generative} to learn appropriate distribution in a data-driven manner.

Specifically, we model a skeleton action `attacker'  as a function $\mathcal{A}$ such that $G'=\mathcal{A}(G^0)$. In the common GAN setup, the attacker can be interpreted as a generator of perturbed skeletons (see Fig.~\ref{fig:teaser}). We set up a binary classification network as the discriminator $\mathcal{D}$. The discriminator accepts either the natural graph $\tilde{G}$ or the perturbed graph $G'$ as its input, and predicts the probability that the input graph came from $\mathcal{P}$. The $\tilde{G}$ and $G'$ are kept `unpaired', implying $\tilde{G}$ and $G^0$ are different graphs sampled from the distribution $\mathcal{P}$. 
To formulate the adversarial learning process, we leverage the least squares  objective~\cite{mao2017least} to train the attacker $\mathcal{A}$ and the discriminator $\mathcal{D}$ using the following loss functions:
\begin{equation}
\label{func:loss_adv_g}
    \mathcal{L}_{adv}(\mathcal{A})=\mathbb{E}_{G'\sim \mathcal{P'}}[(\mathcal{D}(G')-1)^2],
\end{equation}
\begin{equation}
\label{func:loss_adv_d}
    \mathcal{L}_{adv}(\mathcal{D})=\mathbb{E}_{\tilde{G} \sim \mathcal{P}}[(\mathcal{D}(\tilde{G})-1)^2] + \mathbb{E}_{G'\sim \mathcal{P'}}[\mathcal{D}(G')^2].
\end{equation}
During training, $\mathcal{A}$ and $\mathcal{D}$ are optimized jointly. We discuss the related implementation details  in Section~\ref{sec:MD}.

\subsubsection{Localized joint perturbation}
\label{sec:local_pert}
Unlike the pixel space of images, a skeleton action graph has highly discrete structure along both spatial and temporal dimensions. 
This discreteness poses unconventional challenges for adversarial attacks in this domain. 
Nevertheless, it also gives rise to interesting investigation directions. 
For instance, it is intriguing to devise a \textit{localized} adversarial attack which fools the model by perturbing only a particular part of the skeleton graph.  If we closely observe a skeleton action, it is clear that different body joints contribute differently to our perception of actions. 
Additionally, most of the human actions are recognizable by the motion patterns associated with the dominant body parts, e.g.~arms and legs. Such observations make localized perturbations particularly relevant to the skeleton data.  


Localized joint perturbations allow for less variations in the overall skeleton, which is beneficial for imperceptibility. They also provide a controlled injection of regional modification to the target skeleton action. To allow  that, we define a subset of joints within a skeleton as the attack region. Only the joints in that region are modified for localized perturbations. Consequently, all the constraints in Section~\ref{sec:BG_imp} still hold for the attack.


\section{Attacker Implementation}
\label{sec:MD}

\subsection{One-Step Attack}
\label{sec:MD_fgsm}
First, we adopt the Fast Gradient Sign Method (FGSM)~\cite{goodfellow2014explaining} as a primitive attack  to create skeleton perturbation $V'$ in a single step. This adoption allows us to put our attack in a better context for the active community in the direction of adversarial attacks. For the FGSM based attack in our setup, the perturbation computation can be expressed as:
\begin{equation}
\label{func:fgsm}
    V' = V^0 + \epsilon~ \mathrm{sign}({\nabla}_{V^0} \mathcal{L}(\mathcal{F}_{\theta}(V^0,E^0), c_{gt}))
\end{equation}
where $\mathcal{F}_{\theta}$ denotes trained ST-GCN~\cite{yan2018spatial} model, $\mathcal{L}$ is the cross-entropy loss for action recognition, and ${\nabla}_{V^0}$ is the derivative operation that computes the gradient of ST-GCN loss w.r.t.~$V^0$, given the current model parameters $\theta$ and the ground truth action label $c_{gt}$. The sign of gradient is scaled with a  parameter $\epsilon$, and  added to the original graph $V^0$. The FGSM-based attack is computationally efficient as it takes only a single step in the direction of increasing the recognition loss of the target model.  

The basic FGSM attack does not specify the label for the misclassified action, and therefore is a `non-targeted' attack. If we specify a particular label for $c_{gt}$ in Eq.~\ref{func:fgsm}, and subtract the  gradient's sign from the original graph $V^0$ (instead of adding it, as in Eq.~\ref{func:fgsm}) the resulting attack becomes a targeted attack~\cite{kurakin2016adversarial-a} that is likely to change the predicted  label of the considered action to a pre-specified label. 


\subsection{Iterative Attack}
\label{sec:MD_ciasa}
The FGSM attack takes a single step over the model cost surface to increase the loss for the given input. An intuitive extension of this notion is to iteratively take multiple steps while adjusting the step direction~\cite{kurakin2016adversarial-b}. 
For the iterative attack, we also adopt the same technique for the skeleton graph input. However, here we focus on targeted attacks. This is because (a) targeted attacks are more interesting for the real-world applications, and (b) non-targeted attacks can essentially be considered a degenerate case of the targeted attack, where the target label is chosen at random. Hence, an effective targeted attack already ensures non-targeted model  fooling.  To implement, we specify the desired target class and take multiple steps in the direction of decreasing the prediction loss of the model for the target class.

We implement the iterative targeted attack while enforcing the constraints discussed in Section~\ref{sec:BG_imp}. The resulting algorithm is termed as Constrained Iterative Attack for Skeleton Actions (CIASA). At the core of CIASA is the following iterative process: 
\begin{equation}
\label{func:im}
    V'_{0} = V^0;~V'_{N+1} = \mathcal C(V'_{N} - \alpha~ ({\nabla}_{V'_{N}}\mathcal{L}_\mathrm{CIASA}(V'_{N}, c_{target}))),    
\end{equation}
{
At each iteration, $V'_{N}$ is adjusted towards the direction of minimizing the overall CIASA loss $\mathcal{L}_\mathrm{CIASA}$ using a step size $\alpha$. This is equivalent to a gradient descent iteration with $\alpha$ as the learning rate, where the skeleton graph $V'_{N}$ is treated as the model parameter. Hence, we directly exploit the Adam  Optimizer~\cite{kingma2014adam} in the PyTorch library\footnote{https://pytorch.org/} for this computation. The operation $\mathcal C(.)$ in Eq.~\ref{func:im} truncates and realigns the values in its argument with pre-set conditions, explained below.
}

In Eq.~\ref{func:im}, the overall CIASA loss $\mathcal{L}_\mathrm{CIASA}$ consists of the following components:
\begin{equation}
\label{func:loss}
    \mathcal{L}_\mathrm{CIASA} = \mathcal{L}_{pred} +\lambda(\mathcal{L}_{smooth}+\mathcal{L}_{adv}))    
\end{equation}
where $\mathcal{L}_{pred}$ is the cross-entropy loss of the model prediction on $V'$ for the desired target class $c_{target}$. $\mathcal{L}_{smooth}$ is the temporal smoothness loss calculated according to Eq.~\ref{func:loss_smooth}. GAN regularization loss $\mathcal{L}_{adv}$ is a combination of $\mathcal{L}_{adv}(\mathcal{A})$ and $\mathcal{L}_{adv}(\mathcal{D})$ given in Eq.~\ref{func:loss_adv_g} and Eq.~\ref{func:loss_adv_d}. $\lambda$ is a weighting hyper-parameter to balance the individual loss components.

\begin{algorithm}[t]
\caption{Constrained iterative attacker $\mathcal{A}$ to fool skeleton-base action recognition.}
\label{alg:ciasa}
\textbf{Input}: Original graph nodes $V^0\in \mathbb{R}^{3\times N\times T}$, trained ST-GCN model $\mathcal{F}_{\theta}()$, desired target class $c_{target}$, perturbation clipping factor $\epsilon$, max\_iter=$M$, learning rate $\alpha$

\textbf{Output}: Perturbed graph nodes $V'\in \mathbb{R}^{3\times N\times T}$.
\begin{spacing}{1.2}
\begin{algorithmic}[1]
\State set initial $V'=V^0$ 
\While {$i<M$}
\State feed forward $Z=\mathcal{F}_{\theta}(V')$
\State $\mathcal{L}_{pred}=\mathrm{CrossEntroyLoss}(Z,c_{target})$
\State $\mathcal{L}_{smooth}=\frac{1}{T-1}\sum^T_{t=2}\ddot{f'_t}$
\State \label{alg:loss_adv} $\mathcal{L}_{adv}=(\mathcal{D}_{\omega}(V')-1)^2 + (\mathcal{D}_{\omega}(\tilde{V})-1)^2 + \mathcal{D}_{\omega}(V')^2$
\State $\mathcal{L}_\mathrm{CIASA}=\mathcal{L}_{pred}+\lambda(\mathcal{L}_{smooth}+\mathcal{L}_{adv}$)
\State \label{alg:grad} $(\mathcal{L}_\mathrm{CIASA}).\mathrm{Backward}() \Rightarrow gradients $
\State $V',\omega=\mathrm{AdamOptimizer}([V', \omega], ~gradients)$
\If {$|{V'-V^0}| > \epsilon$}
\State $V' = \mathrm{Clip}(V') \sim [V^0-\epsilon, V^0+\epsilon])$
\EndIf
\State Skeleton realignment $V'=\mathrm{SSR(V')}$ 
\State $i=i+1$
\EndWhile
\State \Return $V'$
\end{algorithmic}
\end{spacing}
\end{algorithm}

 Implementing the process identified by Eq.~\ref{func:im} produces the perturbed skeleton $V'$ that fools the model into miscalssifying the original action as $c_{target}$, while complying to the spatio-temporal constraints derived  in the previous Sections. The pseudo-code of implementing the process of Eq.~\ref{func:im} as CIASA is presented in Algorithm~\ref{alg:ciasa}.
The algorithm starts with a forward-pass of $V'$ through the target model $\mathcal{F}_{\theta}()$, i.e.~ST-GCN. The respective losses are then computed to form the overall CIASA loss $\mathcal{L}_\mathrm{CIASA}$. At line~\ref{alg:loss_adv}, $\mathcal{L}_{adv}$ is computed as the accumulation of the losses defined in Eq.~\ref{func:loss_adv_g} and Eq.~\ref{func:loss_adv_d}. Here, we replace $G$ with $V$ based on the algorithm context. $\mathcal{D}_{\omega}$ denotes the discriminator network which is parameterized by $\omega$. Note that, the real data $\tilde{V}$ and the perturbed data ${V'}$ are unpaired, as discussed in Sect.~\ref{sec:Ant}. On line~\ref{alg:grad}, the gradient information is obtained through the back propagation operation denoted as `$.\mathrm{Backward}()$'. We employ the Adam Optimizer~\cite{kingma2014adam} to update the  skeleton joints ${V'}$ and the discriminator parameters  $\omega$. Clipping operation is then applied to truncate ${V'}$ to pre-set ranges.
In our case, the scaling factor $\epsilon$ restricts the $\ell_{\infty}$-norm of the perturbation at graph nodes. For global clipping, $\epsilon\in \mathbb{R}$ is a scalar value that results in equal clipping on all joints. For the hierarchical clipping, $\epsilon\in \mathbb{R}^N$ defines different clipping strengths for different joint. The clipping imposes the \textit{joint variation constraint} over the perturbations. To impose the \textit{bone length constraint}, Spatial Skeleton Realignment (SSR) is proposed to realign the skeleton bones within the clipped ${V'}$ according to the original bone lengths. Note that the operations of clipping and realignment constitute the function $\mathcal C(.)$ shown in Eq.~\ref{func:im}. We empirically set the weight factor $\lambda$ as 10, and the base learning rate for the Adam Optimizer $\alpha$ as 0.01. 
Below we discuss the implementation of SSR and discriminator network $\mathcal{D}$.

\subsubsection{Spatial Skeleton Realignment}
We propose Spatial Skeleton Realignment (SSR) to preserve the \textit{bone length constraint} as we perturb the skeleton graph. SSR is executed at each iteration after $V'$ is updated and clipped in order to realign every perturbed skeleton frame based on the original bone lengths. Specifically, for every updated skeleton joint $v'_j$, we find its parent joint $v'_i$ along the intra-body edge $E^S$. The bone between joints $i$ and $j$ is defined as a vector $b'_{ij}=v'_j - v'_i$. Then, we modify the joint $v'_j$ along the vector direction $\overline{b'_{ij}}$ to meet the constraint in Eq~\ref{func:bl}. The modification applied to $v'_j$ is also applied to all of its children/grandchildren joints. To complete the SSR, the above process starts from the root joint and propagates through the whole skeleton.

\subsubsection{GAN Regularization}
To enforce the anthropomorphic  plausibility of the perturbed skeleton action, the adversarial regularization term $\mathcal{L}_{adv}$ is optimized jointly with the other attack objectives. Taking per-frame skeleton feature map, say $X$ as the input, a discriminator network $\mathcal{D}$ is trained to classify the skeleton as \textit{fake} or \textit{real} (i.e. perturbed v.s original), while the attacker $\mathcal{A}$ is competing with $\mathcal{D}$ to increase the probability of the perturbed skeleton being classified as \textit{real}.

We leverage the angles between skeleton bones to construct the feature map $X$. For a pair of bones $b_{ij}$ and $b_{uv}$, the corresponding element in the feature map is defined as the cosine distance between the bones as:
\begin{equation}
    x_{ij-uv} = \frac{b_{ij} \cdot b_{uv}}{\|b_{ij}\| \|b_{uv}\|}
\end{equation}
We select a group of major bones to construct the feature map $X$, while insignificant bones of fingers and toes are excluded to avoid unnecessary noise. The resulting feature map has dimension $X\in \mathbb{R}^{C,H,W}$, where $C=1$, and $H=W$ equals to the number of selected bones. 
We model $\mathcal{D}$ as a binary classification network that consists of two convolution layers and one fully-connected layer. The convolution kernel size is 3, and the number of channels produced by the convolution is 32. $\mathcal{D}$ outputs values in the range [0, 1], signifying  the probability that $X$ is a real sample.

\section{Experiments}
\label{sec:EXP}
Below we evaluate the effectiveness of the  proposed attack for skeleton-based action recognition. We examine different attack modes on standard skeleton action datasets. We also demonstrate the transferability of attack and explore generalization of the computed adversarial perturbations beyond the skeleton data  modality. Lastly, an ablation study is provided to highlight the contributions of various constraints to the overall fooling rate achieved by the proposed attack.

\subsection{Dataset and Evaluation Metric}
\textbf{NTU RGB+D:} NTU RGB+D Human Activity Dataset is collected with Kinect v2 camera and includes 56,880 action samples. Each action has RGB, depth, skeleton and infra-red data associated with it. However, we are only concerned with the skeleton data in this work. For the skeleton-based action recognition with ST-GCN, we follow the standard protocols defined in \cite{shahroudy2016ntu}, i.e. cross-subject and cross-view recognition. Accordingly, two different ST-GCN models are used in our experiments, one for each protocol. We denote these models as NTU$_\mathrm{XS}$ and NTU$_\mathrm{XV}$ for cross-subject and cross-view recognition. While the original dataset is split into training and testing sets, we only manipulate the testing set, as no separate training data is required for the attack. 


\textbf{Kinetics:} Kinetics dataset~\cite{kay2017kinetics} is a large unconstrained action dataset with 400 action classes. For  skeleton-based action recognition using this data, the original  ST-GCN~\cite{yan2018spatial} first uses OpenPose~\cite{cao2017realtime,cao2018openpose} to estimate 2D skeletons with 18 body joints.
Then, the estimation confidence `c' for every joint is concatenated to its 2D coordinates (x, y) to form a tuple (x,y,c). The tuples for all joints in a skeleton are collectively considered as an input sample by the ST-GCN model. For the adversarial attack, we mask the channel of confidence values and only perturb the (x,y) components for the Kinetics dataset. 

\textbf{Evaluation metric:}
The evaluation metric used to evaluate the success of adversarial attacks is known as   \textit{fooling rate}~\cite{akhtar2018threat}. It indicates the percentage of data samples over which the model changes its predicted label after the samples have been adversarially perturbed. In the adversarial attacks literature, this is the most commonly used metric to evaluate an attack's performance~\cite{akhtar2018threat}. In the case of targeted attacks, it determines the percentage of the samples successfully misclassified as the target label after the attack.



\subsection{Non-targeted Attack}
\begin{figure}[t]
\centering
\includegraphics[width=0.49\textwidth]{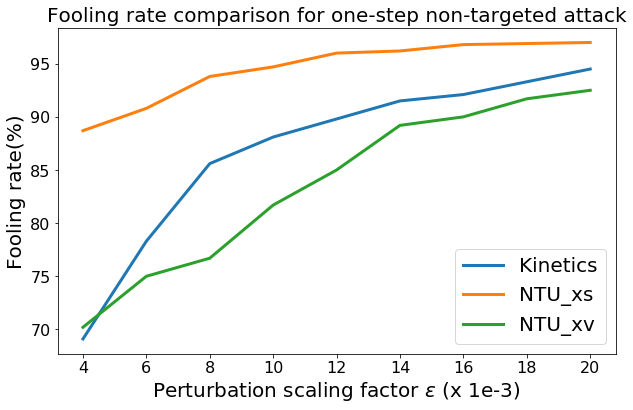}
\caption{Fooling rates (\%) achieved by one-step non-targeted attack with different perturbation scaling factors  for NTU and Kinetics datasets. Both cross-subject NTU$_\mathrm{XS}$ and cross-view NTU$_\mathrm{XV}$ protocols are considered for the NTU dataset.}
\label{fig:fgsm_epsilon_compare}
\end{figure}

Since this is the first work in the direction of attacking skeleton-based action recognition, it is important to put our attacking technique into  perspective. Hence, we first conduct a simpler non-targeted attack on NTU and Kinetics datasets using the one-step attack discussed in Section~\ref{sec:MD_fgsm}, Eq.~(\ref{func:fgsm}). We compute the fooling rates for both datasets under different values of the perturbation  scaling factor $\epsilon$.
Both cross-view and cross-subject protocols were considered in this experiment for the NTU dataset. The fooling rates achieved with the one-step method for various $\epsilon$ values are summarized in Fig.~\ref{fig:fgsm_epsilon_compare}.
As can be seen, the non-targeted fooling is reasonably successful under the proposed formulation of the problem for skeleton-based action recognition.
The fooling rates for all protocols remain higher than 90\% once the $\epsilon$ value reaches 0.02. This is still a reasonably small perturbation value that is equivalent to one twentieth of the average skeleton height. 

\begin{figure*}[t]
\centering
\includegraphics[width=1\textwidth]{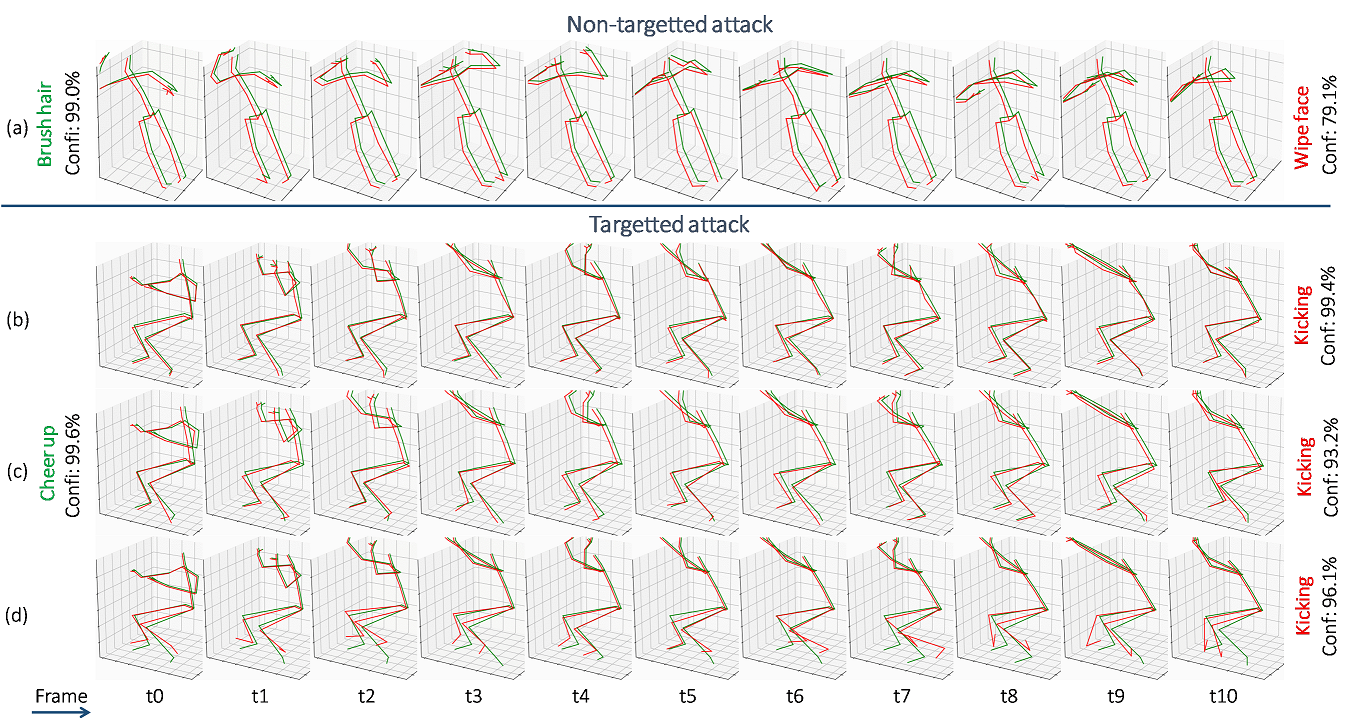}
\caption{TOP: (row a) One-step attack with $\epsilon$ = 0.02 is shown where ``brush hair'' action is misclassified as ``wipe face''. BOTTOM: CIASA targeted attack in different modes are shown. (row b) The \textit{basic} mode that perturbs all joints with $\epsilon$ = 0.01. (row c) The \textit{localized}  mode with only two legs allowed to be perturbed. Global clipping is applied with $\epsilon$ = 0.08. (d) The \textit{advanced} mode where the same two legs are perturbed with hierarchical clipping. The attacks in all modes successfully fool the recognition model with confidences higher than 90\%. The temporal dimension evolves from left to right.}
\label{fig:ciasa_demo}
\end{figure*}

To visualize perturbed skeletons, Fig.~\ref{fig:ciasa_demo}(a) shows a successful attack on NTU dataset for cross-view fooling. The original and perturbed skeletons are plotted with green and red colors respectively.
Note that, in this illustration and the examples to follow, we provide a positional offset between different skeletons for better visualization.
For the shown sequence of skeleton frames, the original label is `Brush hair', that is predicted as `Wipe face' after the attack is performed. The temporal dimension evolves from left to right. Ignoring the positional offset, it is easy to see that the perturbation generally remains hard to perceive in the skeleton.



\subsection{Targeted Attack}

We use the proposed CIASA attacker explained in Section~\ref{sec:MD_ciasa}, Alg.~\ref{alg:ciasa} to conduct targeted attacks on both  NTU and Kinetics datasets. We specify the least-likely action prediction of the ST-GCN models as the target label $c_{target}$ as described in Eq.~\ref{func:im},
 implying that the most challenging misclassification target is chosen to launch attacks. CIASA is configured to launch attacks in three modes; namely, \textit{basic} mode, \textit{localized} mode, and \textit{advanced} mode. Below we discuss these modes along the experimental results.

Figure~\ref{fig:ciasa_demo}(b) shows an example of CIASA attack in the \textit{basic mode}. We apply the global clipping discussed in Section~\ref{sec:BG_imp} in this attack mode, where all the skeleton joints are perturbed with the same scaling factor $\epsilon$ = 0.02. With this setting, the original action of `Cheer up' in Figure~\ref{fig:ciasa_demo}(b) is misinterpreted  as `Kicking' with confidence score 99.4\%. 
In the basic mode, the comparison of fooling rates with different $\epsilon$ values for the two benchmark datasets are summarized in Table~\ref{tab:targeted_attack_result}. Firstly, the results demonstrate successful fooling even for very low $\epsilon$ values. Secondly, it is noteworthy that for similar $\epsilon$ values, higher fooling rates are generally achieved by CIASA for targeted fooling as compared to the non-targeted fooling of the one step method in Fig.~\ref{fig:fgsm_epsilon_compare}. This demonstrates the strength of CIASA as a targeted attack. 
In our experiments, we observed that the least-likely label of ST-GCN model remains similar for multiple  actions. Whereas the presented results do not diversify the target labels of such actions to strictly follow the evaluation protocol, it is possible to manually do so. Loosening the evaluation criterion on these lines will further improve the fooling rate of CIASA. 



\begin{table}[t]
\centering
\caption{Fooling rates (\%) achieved by CIASA targeted attack (basic mode) with different global clipping strength $\epsilon$ for NTU and Kinetics datasets. Both cross-subject NTU$_\mathrm{XS}$ and cross-view NTU$_\mathrm{XV}$ protocols are considered for the NTU dataset.}
\label{tab:targeted_attack_result}
\begin{tabular}{l|ccccc}
\hline\noalign{\smallskip}
$\epsilon$~($\times$ 1e-3) & 4 & 6 & 8 & 10 & 12\\
\noalign{\smallskip}\hline\noalign{\smallskip}
Kinetics & 82.5 & 92.5 & 96.5 & 97.5 & 99.3 \\ 
NTU$_\mathrm{XS}$ & 89.4 & 96.6 & 98.7 & 99.2 & 99.8 \\ 
NTU$_\mathrm{XV}$ & 78.2 & 85.5 & 93.3 & 98.9 & 99.6 \\
\noalign{\smallskip}\hline
\end{tabular}
\end{table}

In Fig.~\ref{fig:ciasa_demo}(c), we shows an example of CIASA attack in the \textit{localized mode}, where the localized joint perturbation discussed in Section~\ref{sec:local_pert} is applied. 
In this example, two legs of skeleton are set to be the attack regions, which allow 8 active joints for perturbations. The remaining joints are unaffected  by the computed perturbations. Compared to the basic mode, fewer joints contribute to the overall perturbation in the localized mode. To compensate for the reduced number of active joints, we loose the perturbation scaling factor and set $\epsilon$ to 0.08 for this experiment. For the shown example, CIASA achieves fooling with 93.2\% confidence for this mode, which is still competitive to the 99.4\% confidence in the basic mode.

\begin{figure}[t]
\centering
\includegraphics[width=0.45\textwidth]{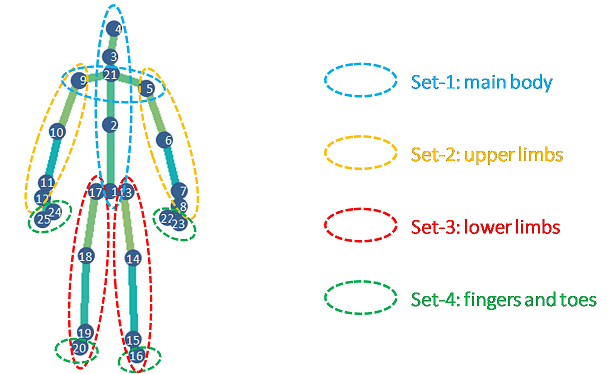}
\caption{The skeleton of NTU dataset is spitted into 4 attack regions, each of which is activated to apply CIASA localized attacks. Every attack region consists of roughly the same number of joints.}
\label{fig:skeleton_sets}
\end{figure}

To further evaluate the localized mode of CIASA with different attack regions, we split the skeleton joints into 4 sets, as illustrated in Fig.~\ref{fig:skeleton_sets}. Then, we conduct CIASA localized attack on NTU dataset for the 4  sets separately. Global clipping is applied for these experiments with the scaling factor $\epsilon$ = 0.04. The chosen value of $\epsilon$ is intentionally kept lower than that in Fig.~\ref{fig:ciasa_demo}(c) because we focus on analysing the fooling prowess of  different attack regions instead of simply  achieving high fooling rates for all the regions. The results of our experiments are  summarized in Table~\ref{tab:sparsity_attack_result}. 
It is clear that the CIASA localized attack achieves impressive  fooling rates by perturbing only a small set of joints within the skeleton. In addition, different sets of active joints affect the fooling performance differently. In Table~\ref{tab:sparsity_attack_result}, set-1 and set-2 achieve higher fooling rates than the other two sets. This can be explained by the observation that many dominant movements in the NTU dataset occur  at the upper part of human body.


\begin{table}[t]
\centering
\caption{Fooling rate(\%) achieved by CIASA targeted attack (localized mode) with different attack regions on NTU dataset. Both cross-subject and cross-view protocols are evaluated. Global clipping strength is set to  $\epsilon$ = 0.04.}
\label{tab:sparsity_attack_result}
\begin{tabular}{l|cccc}
\hline\noalign{\smallskip}
Attack region & set-1 & set-2 & set-3 & set-4 \\
\noalign{\smallskip}\hline\noalign{\smallskip}
NTU$_\mathrm{XS}$ & 90.8 & 93.3 & 61.3 & 83.3 \\ 
NTU$_\mathrm{XV}$ & 85.2 & 91.7 & 60.0 & 81.7 \\
\noalign{\smallskip}\hline
\end{tabular}
\end{table}

We also extend the  localized mode of CIASA to an  \textit{advanced mode} by replacing the  global clipping by  hierarchical clipping discussed in Section~\ref{sec:BG_imp}. In that case, the scalar clipping value $\epsilon$ is replaced by $\epsilon\in \mathbb{R}^N$, where N is the number of active joints to be perturbed. Here, we allow various active joints to change in pre-defined ranges by using differentiated clipping values. One strategy to differentiate the clipping strength is applying incremental $\epsilon$ variables from parent joints to their children joints, based on the observation that children joints normally move in larger ranges than their parents. Figure~\ref{fig:ciasa_demo}(d) illustrates an example of successful advanced attack on NTU dataset with two legs activated for the attack. The $\epsilon$ variables are set to 0.01, 0.05, 0.15, 0.25 for the joint \textit{hips}, \textit{knees}, \textit{ankles}, and \textit{feet}, respectively. Note that we intentionally amplify the perturbation ranges at certain joints such as \textit{ankles} and \textit{feet}, which results in noticeable perturbations at the attack region. We will justify the intuition behind this differential clipping  in the paragraphs to follow.
For now, notice that in Fig.~\ref{fig:ciasa_demo}(d), the original label of `Cheer up' is misclassified as `Kicking' with a confidence score 96.1\% with the advanced  attack.

Although the CIASA attack in \textit{advanced mode} apparently sacrifices the visual imperceptibility of the perturbation, it is able to maintain  the ``semantic imperceptibility'' for   the perturbed skeleton. We corroborate this claim with the following   observations.
First, in Fig.~\ref{fig:ciasa_demo}(d), the dominant body movements for  `Cheer up' action mainly occur in the upper part of the skeleton, while the fooling is conducted  by perturbing the lower body to which less attention is paid for this action.  Consequently, the attack does not incur significant perceptual attention in the first place. Furthermore, due to the spatio-temporal constraints with CIASA attacks, the injected perturbation patterns remain smooth and natural. This further reduces the attack suspicions, as compared to any small but unnatural perturbations, e.g.~shakiness around the joints.

Further to the above discussion, the perturbations generated in the advanced mode can not only fool the recognition model in skeleton spaces, but can also be imitated and reproduced in the Physical world. Imagine an end-to-end skeleton-based action recognition system using a monocular camera as its input sensor. For that, RGB images taken from the Physical world are first converted to skeleton frames, which are then passed through  the skeleton-based action recognition model. For this typical  pipeline, it may be  inconvenient to interfere with the intermediate skeleton data for the attacking purpose. However, the adversarial perturbations can be injected into the input RGB data by performing an action in front of the camera while imitating the perturbation patterns with selective body parts.  The advanced mode of CIASA allows the discovery of perturbation patterns for such attacks. 
This is elaborated further in Section~\ref{sec:tran_modality} with relevant context.



\subsection{Transferability of Attack}

We examine the transferability of the proposed CIASA attack from two perspectives. First, we evaluate the cross-model transferability of the generated perturbations. Concretely, we attack a skeleton action recognition model A to generate perturbed skeletons. Then, we  predict the label of the perturbed skeletons using model B and examine the fooling rate for model B. We respectively chose ST-GCN and 2s-AGCN~\cite{shi2019two} as model A and B in our experiments.

Second, we analyze the cross-modality transferability of CIASA attack. i.e. we generate perturbations for one data modality and test their fooling capability in another data modality. We formulate this task as transferring perturbations from skeleton data to RGB data, as RGB cameras are widely used as input sensors for the real world systems. For the cross-modality test, we generate perturbed skeletons by attacking the ST-GCN. Then, those skeletons are converted to RGB actions using a graphics rendering pipeline. To examine whether the adversarial information can be preserved during the conversion, we predict the label of RGB actions under the usual skeleton-based action recognition pipeline for the ST-GCN.



\subsubsection{Cross-Model Transferability}
\label{sec:tran_model}
The 2s-AGCN~\cite{shi2019two} is a two-stream adaptive graph convolutional network for skeleton-based action recognition. This network is significantly different from the ST-GCN~\cite{yan2018spatial} as it models a learnable topology of the skeleton graph. In addition to the joint locations, 2s-AGCN also models the bone directions, which results in a two-stream network structure.

We first generate perturbed skeleton actions based on ST-GCN model. The \textit{basic mode} of CIASA with global clipping is employed, where the perturbation scaling factor $\epsilon$ is empirically set to 0.012. The cross-view protocol of NTU dataset is adopted to create perturbed skeletons, which are then evaluated by 2s-AGCN models. We compare the change of recognition accuracy before and after the attack, and record the fooling rates for three different configurations of the 2s-AGCN, i.e. joint only (Js-AGCN), bone only (Bs-AGCN) and ensemble (2s-AGCN). The results in Table~\ref{tab:cross_model_attack} show  that the perturbations generated with ST-GCN significantly degrades the recognition performance of 2s-AGCN. 
This demonstrate that the proposed CIASA attacker is able to generalize well on `unseen' action recognition models.

\begin{table}[t]
\centering
\caption{Comparison of cross-model recognition accuracy (\%) and fooling rate (\%) on three configurations of 2s-AGCN for cross-view NTU protocol. `Original Accuracy' is on clean data. `Attacked Accuracy' is  on perturbed data.}
\label{tab:cross_model_attack}
\begin{tabular}{l|ccc}
\hline\noalign{\smallskip}
Model & Js-AGCN & Bs-AGCN & 2s-AGCN \\
\noalign{\smallskip}\hline\noalign{\smallskip}
Original Accuracy & 93.7 & 93.2 & 95.1 \\ 
Attacked Accuracy & 13.5 & 6.8 & 11.8 \\
\hline\noalign{\smallskip}
Fooling rate (\%) & 86.1 & 93.1 & 88.4 \\
\noalign{\smallskip}\hline
\end{tabular}
\end{table}

\begin{figure*}[t]
\centering
\includegraphics[width=1\textwidth]{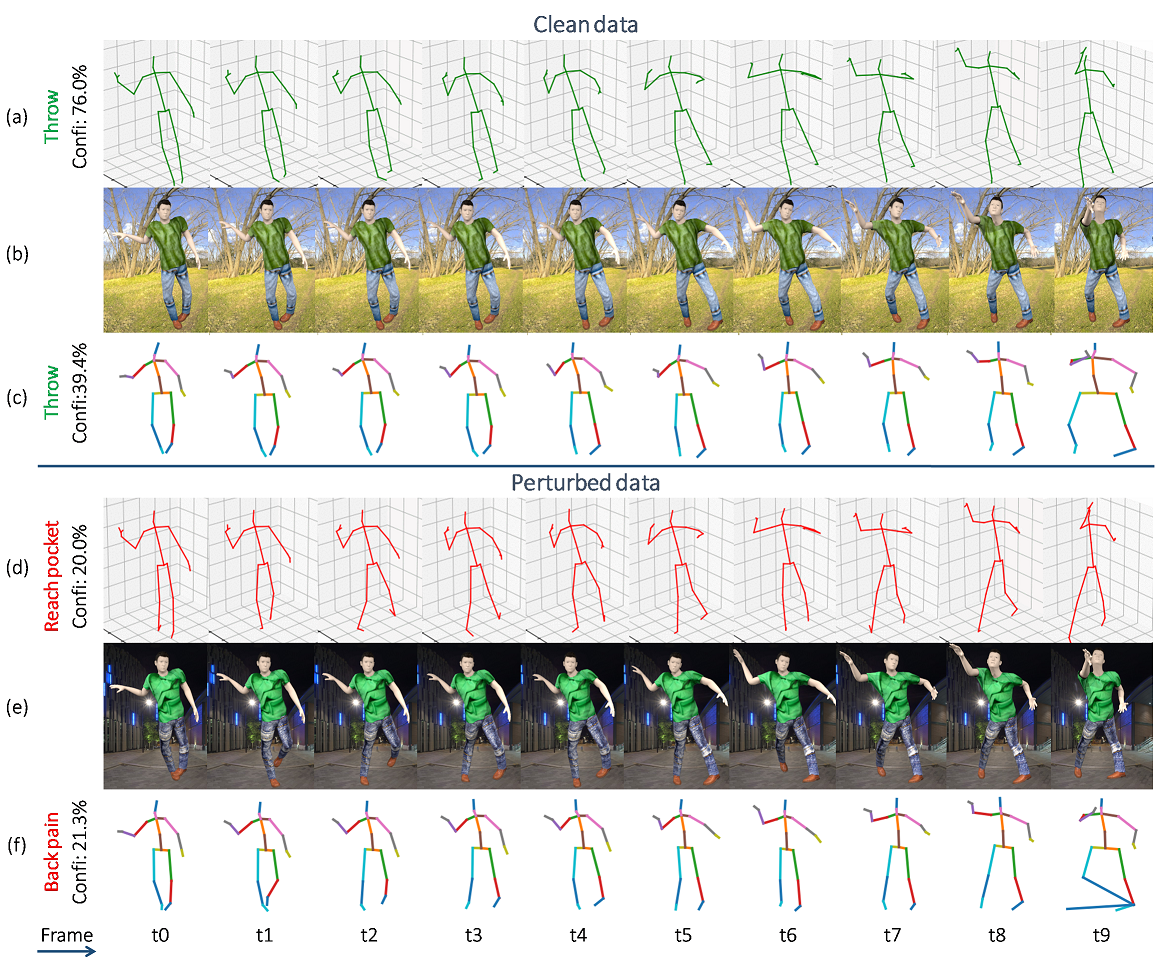}
\caption{TOP: Clean data of different modalities. (a) Original skeleton sequences. (b) RGB video rendered from the original sequence. (c) Recovered 3D pose sequence extracted from (b) using VNect~\cite{mehta2017vnect}. BOTTOM: Perturbed data of different modalities. (d) Perturbed skeleton sequences created with the advanced mode of CIASA. (e) RGB video rendered from (d). (f) 3D poses extracted from (e) using VNect~\cite{mehta2017vnect}. 
}
\label{fig:vnect_recognition}
\end{figure*}

\subsubsection{Cross-Modality Transferability}
\label{sec:tran_modality}
To transfer the perturbations from skeleton to RGB space, we adopt a human pose synthesis technique~\cite{liu2017learning} to create RGB actions based on the perturbed skeleton sequences generated with the advanced mode of CIASA. The adopted synthesis pipeline can produce realistic RGB actions with diversified human models, backgrounds, cloth textures and illuminations. Moreover, the temporal dynamics of the underlying action is also reproducible in the synthesized RGB video. We demonstrate successful cross-modality transferability in Fig.~\ref{fig:vnect_recognition}. The rows (a) and (d) are the original and perturbed skeleton sequences respectively. (b) and (e) show the RGB actions generated using~\cite{liu2017learning} with (a) and (d) used as the inputs skeleton sequences.

Firstly, the successful generation of realistic RGB videos in (b) and (e) affirms that the skeleton perturbations generated by CIASA are useful in producing action perturbations in the Physical world beyond the skeleton space. 
Secondly, we observe that the adversarial information remains largely preserved during the cross-modality transfer. In Fig.~\ref{fig:vnect_recognition}, we use  VNect~\cite{mehta2017vnect} as a 3D pose extractor to recover 3D skeletons directly from the synthesized RGB actions. The recovered skeleton sequences are then fed to the trained ST-GCN model for action recognition, mimicking the typical pipeline for the skeleton-based action recognition for RGB sensors.

The VNect-recovered 3D skeletons from clean  and perturbed RGB data are respectively shown in rows (c) and (f) of the figure. As can be seen, the recovered skeletons generally follow the motion patterns encoded in the respective source skeletons. For the clean data, the recovered skeletons in (c) and the source skeletons in (a) are both correctly recognized as `Throw' action. For the perturbed data, the recovered skeleton sequence in (f) has  fooled the ST-GCN into  miscalssifying the action as `Back pain'. Although the fooling is not in the exact least likely class as in row (d), misclassifcation due to CIASA attack for this very challenging scenario is still intriguing. We note that the attack here is naturally degenerating into an un-targeted attack. 


To further scale up the cross-modality experiment, we randomly select 240 skeleton actions for the cross-view protocol of the NTU dataset. Then, we conduct the cross-modality transfer for all those sequences. We only use a subset of the NTU dataset  because of the unreasonable computational time required to render videos for the complete  dataset. Subsequently, we predict action labels with ST-GCN on the VNect-recovered skeleton sequences for both clean and perturbed data.
With this setting, the recognition accuracy is recorded as 53.3\% for the clean data, and 38.9\% for the perturbed data. Compared to the original NTU cross-view accuracy of 88.3\%~\cite{yan2018spatial}, lower performance is observed on the clean data due to inaccurate 3D pose extraction by VNect. Nevertheless, the proposed attack is still able to preserve its adversarial characteristics to further cause a significant accuracy drop in this challenging scenario. 
\begin{figure}[t]
\centering
\includegraphics[width=0.49\textwidth]{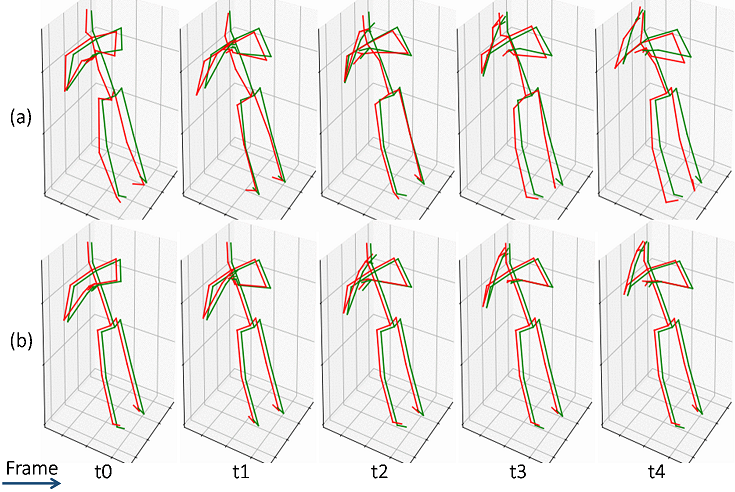}
\caption{Temporal smoothness in CIASA. (a) Perturbed skeleton sequence without temporal smoothness constrains. (b) Perturbed sequence with temporal smoothness constrains. The original and perturbed skeletons are shown in Green and Red colors respectively.}
\label{fig:ablation_smooth}
\end{figure}

\begin{figure}[t]
\centering
\includegraphics[width=0.49\textwidth]{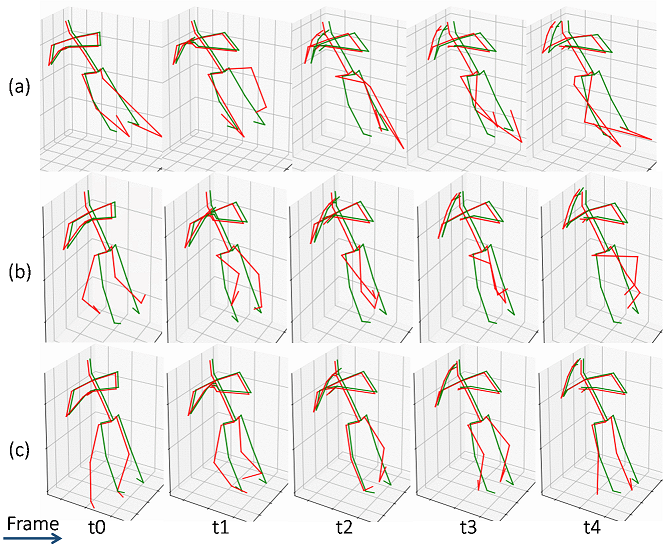}
\caption{Effectiveness of spatial constraints in  CIASA. A localized attack is launched on two legs of the skeleton. (a) No spatial constraints: Pose configuration and bone lengths change randomly. (b) Spatial Skeleton Realignment (SSR): Constrained consistent bone lengths, but  unnatural poses. (c) GAN regularization: Realistic poses that can correspond to the real-world skeleton motions.}
\label{fig:ablation_spatial}
\end{figure}

\subsection{Ablation Study}
For the CIASA attack, we have proposed a set of spatio-temporal constraints to achieve high-quality adversarial perturbations in terms of  both temporal coherence and spatial integrity of the  perturbed skeletons. Here, we provide an ablation study to compare the contributions of these constraints in the overall results.

To enforce  temporal smoothness in the perturbed skeleton sequences, we penalize the joint accelerations between the consecutive skeleton frames. Figure~\ref{fig:ablation_smooth} compares the perturbed skeletons with and without this temporal constrains in the \textit{basic mode} of CIASA, where the original and perturbed skeletons are highlighted with green and red color respectively. It is apparent that the perturbed skeletons in (b) move more smoothly than those in (a) along the temporal dimension. This ascertains the effectiveness of temporal smoothing in our attack. Both perturbations in (a) and (b) successfully fool the recognition model to miscalssify ``Drink action'' as the ``Jump up'' action.

To enforce spatial integrity and anthropomorphic plausibility, we use spatial skeleton realignment (SSR) and GAN regularization. Such spatial constrains are particularly important for the CIASA localized attacks, where only a given subset of the joints is permitted to be changed. Figure~\ref{fig:ablation_spatial} compares the perturbation results with and without the spatial constraints for a localized attack on skeleton legs. Without any spatial constrains, the perturbed skeletons in (a) shows unrealistic pose configurations and arbitrary lengths of bones. With only SSR enabled in (b), lengths of the perturbed bones are more consistent with their original values, however, the resulting poses are still not realistic in terms of plausibility. By adding the GAN regularization, the skeletons in (c) are more realistic. The skeleton sequences in the figure clearly demonstrates the effectiveness of SSR and GAN regularization in our attack.  
All sequences in Fig.~\ref{fig:ablation_spatial} (a), (b) and (c) successfully fool the recognition model in predicting the label ``Drink water'' as ``Jump up''.

\section{Conclusion}
\label{sec:Conc}
We present the first systematic adversarial attack on skeleton-based action recognition.
Unlike the existing attacks that target non-sequential tasks, e.g.~image classification, semantic segmentation and pose estimation, we  attack deep sequential models from a  spatio-temporal perspective. With skeleton-based action recognition model ST-GCN~\cite{yan2018spatial} as the target, we demonstrate its successful fooling by mainly perturbing the joint positions. 
The proposed attack algorithm CIASA imposes spatio-temporal constraints on the adversarial perturbations to produce perturbed skeleton sequences with temporal smoothness, spatial integrity, and anthropomorphic plausibility. The proposed algorithm works in different modes based on the needs of the attack. With the \textit{localized} mode of CIASA, we are able to perturb only a particular set of the body joints to launch localized attack. Such attacks can be used to inject regional perturbations to pre-specified parts of the body, without interfering with the dominant action patterns that are performed by the other joints. Compared to the \textit{basic} mode that perturbs all the joints with global clipping, an \textit{advanced} mode  utilizes localized attacks with hierarchical joint variations to disguises the attack intentions with realistic motion patterns. Our experiments show that the proposed CIASA perturbations generalize well across different recognition models. Moreover, they also have the ability to transfer to RGB video modality under graphics rendering pipeline.  
This indicates that CIASA generated perturbations can allow attackers to mimic semantically imperceptible adversarial patterns in the real world to fool skeleton based action recognition systems. 

\section*{Acknowledgment}
This research is supported by the Australian Research Council (ARC) grant DP190102443. The Tesla K-40 GPU used for this research is donated by the NVIDIA Corporation.

\ifCLASSOPTIONcaptionsoff
  \newpage
\fi



\bibliographystyle{IEEEtran}
\bibliography{IEEEabrv,skeleton_attack}

\balance
\end{document}